\documentclass{article}

\usepackage[nonatbib, final]{nips_2017}

\usepackage[utf8]{inputenc} 
\usepackage[T1]{fontenc}    
\usepackage{hyperref}       
\usepackage{url}            
\usepackage{booktabs}       
\usepackage{amsfonts}       
\usepackage{nicefrac}       
\usepackage{microtype}      

\usepackage{amsmath}
\usepackage{graphicx}
\usepackage{multirow}
\pdfoutput=1

\title{Gradual Tuning: a better way of Fine Tuning the parameters of a Deep Neural Network}

\author{
  Guglielmo Montone\\
  Laboratoire Psychologie de la Perception\\
  Universit\'e Paris Descartes, Paris\\
  \texttt{montone.guglielmo@gmail.com} \\
  \And
  J.Kevin O'Regan\\
  Laboratoire Psychologie de la Perception\\
  Universit\'e Paris Descartes, Paris\\
  \texttt{jkevin.oregan@gmail.com} \\
  \AND
  Alexander V. Terekhov\\
  Laboratoire Psychologie de la Perception\\
  Universit\'e Paris Descartes, Paris\\
  and\\
  Redwood Center for Theoretical Neuroscience\\
  University of California, Berkeley\\ 
  \texttt{avterekhov@gmail.com} \\
}

\begin{document}

\maketitle

\begin{abstract}
In this paper we present an alternative strategy for fine-tuning the parameters of a network. We named the technique \textit{Gradual Tuning}. Once trained on a first task, the network is fine-tuned on a second task by modifying a progressively larger set of the network's parameters. We test \textit{Gradual Tuning} on different transfer learning tasks, using networks of different sizes trained with different regularization techniques. 
The result shows that compared to the usual fine tuning, our approach significantly reduces catastrophic forgetting of the initial task, while still retaining comparable if not better performance on the new task.
\end{abstract}

\section{Introduction}
Due to the huge amount of data needed to train a network it is always more common in the Deep Learning community to use Transfer Learning techniques. Such techniques mainly 
consist in training a network on a new task after the same network has been trained in solving one or 
more tasks similar to the new one. Transfer Learning has been proven to be effective in many domains like image 
recognition, natural language processing and reinforcement learning \cite{rusu2016progressive, 
collobert2008unified, terekhov2015knowledge}. Two strategies are particularly common in Transfer Learning \cite{yosinski2014transferable}. In one strategy, referred to as \textit{Fine 
Tuning}, training on a new task involves allowing modification of all the parameters of the network. The second strategy 
instead consists in adding some new parameters to the network, often by adding new layers on top of the old 
network. The network is then trained on the new task by modifying only the new parameters \cite{montone2015usefulness}. While 
\textit{Fine Tuning} is often the more effective of the two techniques, it has the major drawback of 
suffering from Catastrophic Forgetting (CF): training on the new 
task considerably reduces performance on the first task.\\
In this paper we propose a very simple variation of fine tuning that we called \textit{Gradual Tuning}:
during the training of the network on the second task we involve the parameters of the 
network only gradually, first allowing just the output layer of the network to be updated and 
then progressively involving the rest of the parameters. The justification for this
procedure derives from the following simple consideration: the gradient of the error function with respect to one 
of the features of the network (here the word feature refers to the vector of incoming weights of a 
neuron) is a function of the feature itself but also of the way the feature is connected to the other 
features of the network. If the network has good features at level L for 
solving a given task but these features are not appropriately connected together by the upper layer, then 
both the features of layer L and the upper layers will have a gradient different from zero. So good 
features of layer L may be changed just because the upper layer does not combine them in the proper 
way.\\ We tested \textit{Fine Tuning} and \textit{Gradual Tuning} on different tasks. A first set of 
experiments was run on a transfer learning task build on the MNIST dataset \cite{srivastava2013compete}. 
The rest of the experiments were involved more challenging synthetic tasks. These tasks, 
all consisting in binary classification of images involving the concepts of line and angle, were 
developed to test \textit{Fine Tuning} and \textit{Gradual Tuning} in different conditions. In a first 
experiment we compared the two strategies when the transfer is between two very similar tasks or when 
it is between two more different ones. In order to test how much the quality of the transfer and the 
amount of CF was affected by the quality of the initial features we developed a second group of 
experiments. In these experiments the network was at first trained on a group of 5 tasks, then the 
training on a new task was performed by using both \textit{Fine Tuning} and \textit{Gradual Tuning}.\\
The paper is organized as follows: In the next section we will more clearly describe the \textit{Gradual 
Tuning} procedure. In section \ref{sec:Experiments} and \ref{sec:tasks} we will present all the 
experiments we developed and the dataset used. Finally in \ref{sec:result} we will present the results 
of our experiments.

\section{Gradual training procedure}
In this paragraph we will give a precise description of the \textit{Gradual Tuning} procedure that we used in all the experiments presented in this paper.
Let's consider a network trained on a first task (Task-A) that we wish to train on a second task, Task-B. To do so we connect the last 
hidden layer of the network with some new output neurons, randomly initializing the weights of this new 
layer. Then we start the training on the second task. The training with the \textit{Fine Tuning}
procedure simply consists in evaluating the gradient for all the parameters of the network and updating 
them with the usual update rule. In the \textit{Gradual Tuning} procedure we start by allowing 
just the parameters of the output layer to be updated, then, when the performance of the network 
stops improving, we allow the hidden layer closest to the output to be trained as well. We 
continue in this way allowing the next hidden layer to be updated each time the performance of the 
network stops improving. The process stops when all the layers of the network have been modified by the training.

\section{Experiments}
\label{sec:Experiments}
The experiments described in this paper are all meant to examine the amount of CF in transfer learning tasks when a network is trained with the standard \textit{Fine Tuning} procedure as compared to training with the \textit{Gradual Tuning}. All the experiments were developed on FFW networks (the characteristics of the network can be found at the end of this paragraph). The procedure of the first group of experiments is the following. We first train the network on a first task (Task-A). Then the last hidden layer of the trained network, other than the layer that remains connected to the output layer used for Task-A, is additionally connected to a new output layer that will be used for training the network on a second task (Task-B). The weights of this new output layer are randomly initialized. The network is then trained on Task-B, while the performance of the network is evaluated both on Task-A and Task-B. The training of the network on Task-B is done either with the usual \textit{Fine Tuning} or with the \textit{Gradual Tuning} procedure. We compare the two training procedures in the following situations:
\begin{itemize}
\item when Task-A and Task-B are both either simple or more complex;
\item when Task-A and Task-B are similar or different.
\end{itemize}
In the first experiment we did on MNIST, we split MNIST into two datasets, one containing the digits from zero to four and one containing the digits from five to nine.  Task-A and Task-B consisted both in  5-class classification problems.\\
In order to have more challenging tasks for the network we build a series of synthetic binary classification tasks. All the tasks involve the concepts of line, angle and triangle. In each task the network receives an image as input and has to answer a question like: "is there an angle or a triangle?". 
A precise description of each dataset and the relative binary classification task can be found in 
paragraph \ref{sec:tasks}.  We develop such tasks for different reasons. The first is simply that 
such tasks are more complex than MNIST, meaning that the network needs much more data to have good 
performance (less than $10\%$ misclassified). The 
second reason is that we wanted to have both tasks that were very similar to each 
other and tasks that were more different. In this way we would be able to check to what extent similarity of the tasks influences CF: This was our second 
experiment where a network trained on Task-A was then trained on Task-B in two different conditions, with
Task-B more or less similar to Task-A.\\
A third reason for having a set of similar tasks was that we wished to investigate how much CF depends 
on the quality of the features the network learned during Task-A. To test this we developed the following 
experiment. During the first part of the experiment, the network was trained on several tasks (5-tasks) 
at the same time. This was done by connecting the last hidden layer of the network to five different 
output layers. The training was executed at the same time, meaning that in each epoch the batches of 
data relative to the different datasets were regularly alternated. Paragraph \ref{sec:result} 
provides the details of the training procedure. Once the training was over, a new output layer was connected to the last hidden layer of the network and the network was trained on a second 
task (Task-B). During the training the performance of the network was evaluated for the five classification tasks of Task-A and the classification of Task-B.\\
Two kinds of FFW networks were used during the experiments. The first kind has two fully connected hidden layers of 500 units each. The second kind of network has three fully connected layers of 1000 units each. The activation function of the nodes of the hidden layers is a ReLU function ($f(x) = max(0,x)$). The activation function of the output units is a softmax function. In the following we will refer to the two architectures as the \textit{(500-500)} and the \textit{(1000-1000-1000)} network. The error used to train the network was the categorical cross-entropy.\\
We will frequently refer to two different regularization techniques, namely L1 regularization and Dropout. In the case of L1 regularization the sum of the absolute value of the weights of the network were multiplied by a coefficient $\lambda_1 = 10^{-4}$ and added to the training error of the network. In the case of Dropout, the percentage of neurons that were dropped out during the training was $20\%$ for the first hidden layer of the network and $50\%$ for the other layers.

\section{The tasks}
\label{sec:tasks}
\begin{figure}[t]
\centering
\includegraphics[width=0.65\textwidth]{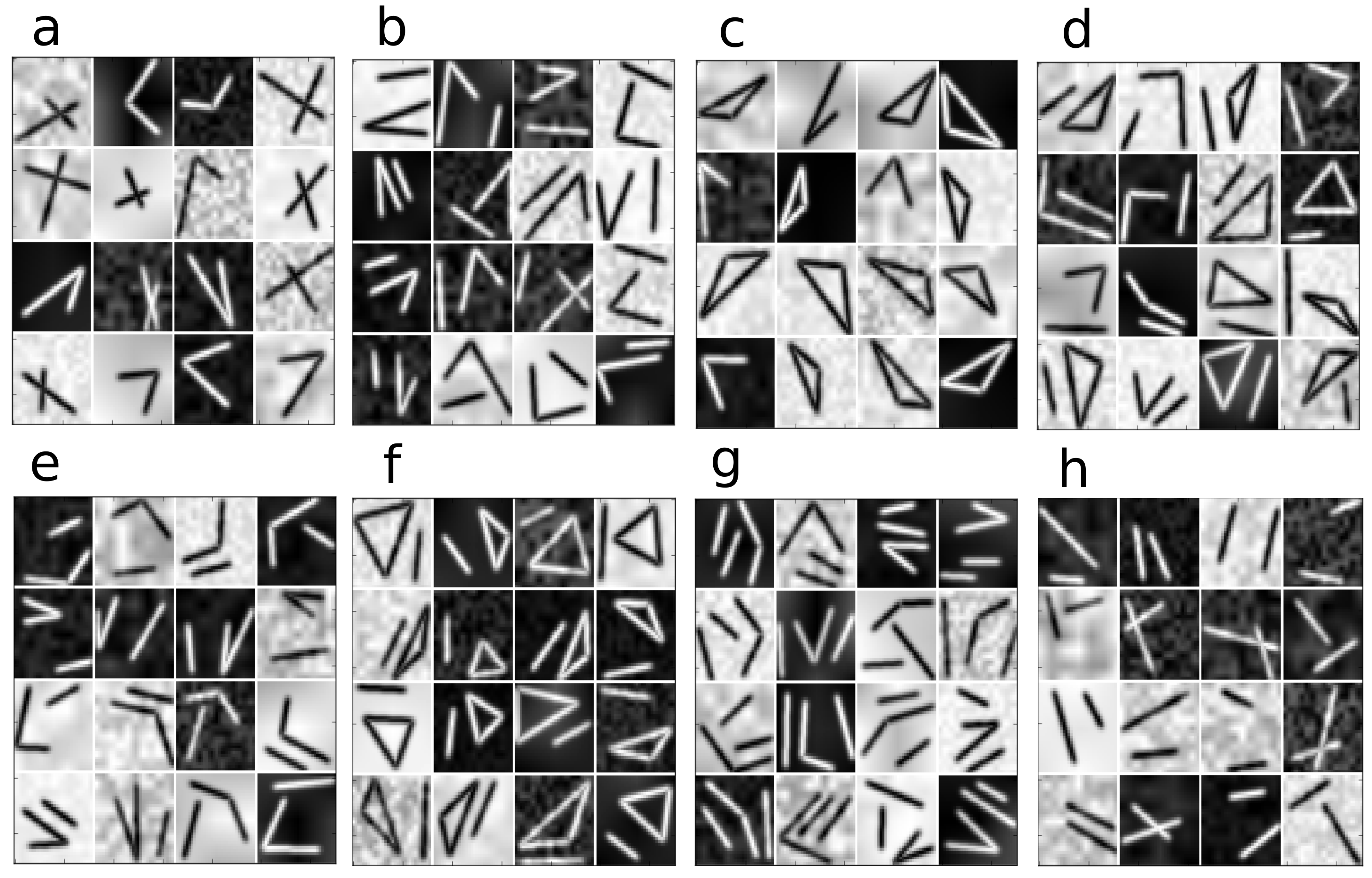}
\caption{In each pictures some samples from the used datasets. (a) \textit{ac} (b) \textit{acl} (c) \textit{at} (d) \textit{atl} (e) \textit{sbl} (f) \textit{sbt} (g) \textit{sb2l} (f) \textit{cnc}}
\label{figure:stimuli}
\end{figure}
The experiments consisted in training a network on classification tasks. In this section we illustrate all the tasks we used. We split the MNIST dataset into two smaller datasets that we named MNIST-04 and MNIST-59. The MNIST-04 dataset contains the MNIST digits between 0 and 4 while the Mnist-59 contains the digits between 5 and 9. Both MNIST-04 and MNIST-59 are composed of approximately 30.000 gray-scale 
images of $28 \times 28$ pixels. Each of the datasets was split into three smaller sets containing 20.000, 
5.000 and 5.000 data points, respectively used as training, validation and test sets.\\
The tasks we built by creating synthetic datasets are all binary classification tasks. In each task the 
stimuli were gray scale images, $32 \times 32$ pixels in size. Each image contained two to four line 
segments, each at least 13 pixels long (30\% of the image diagonal). The segments were white on a dark 
random background or black on a light random background.  Examples of the 8 tasks can be seen in figure \ref{figure:stimuli}):

\vspace{0.5pt}
\textit{ac}: requires classifying the images into those containing an angle (between $20^{\circ}$ and $160^{\circ}$) and a pair of crossing line segments (the crossing point must lie between 20\% and 80\% along each segment’s length).

\vspace{0.5pt}
\textit{acl}: the same as \textit{ac}, but in each picture there is an additional distracting line crossing neither of the other line segments.

\vspace{0.5pt}
\textit{at}: distinguishes between images containing an angle (between $20^{\circ}$ and $160^{\circ}$) and a triangle (with each angle between $20^{\circ}$ and $160^{\circ}$ ). 

\vspace{0.5pt}
\textit{atl}: the same as \textit{at}, but with an additional distracting line, crossing neither of the line segments forming the angle and the triangle.

\vspace{0.5pt}
\textit{sbl}: requires classifying the images into those having blunt (between $100^{\circ}$ and $160^{\circ}$ ) and those having sharp (between $20^{\circ}$ and $80^{\circ}$ ) angles in them. Each image also contains an additional line segment, crossing neither of the line segments forming the angle.

\vspace{0.5pt}
\textit{sbt}: requires classifying the images into those having blunt (between $100^{\circ}$ and $160^{\circ}$ ) and those having sharp (between $20^{\circ}$ and $80^{\circ}$ ) triangles in them, in the picture there is also a distracting line that does not intersect any of the two triangles.

\vspace{0.5pt}
\textit{sb2l}: the same as \textit{sbl}, but has two additional distracting lines, crossing neither of the line segments forming the angle.

\vspace{0.5pt}
\textit{cnc}: distinguishes between a pair of crossing and a pair of non-crossing lines (the crossing
point must lie between 20\% and 80\% of each segment length).

Each stimulus was generated by randomly selecting the parameters describing each figure (length
and orientation of straight segments, amplitude and orientation of the angles) and verifying that all
conditions were satisfied. Four different types of random backgrounds were generated with four
patterns changing at different rates. For our experiments we generated 350.000 stimuli for
each condition. Each dataset was split into three sets containing respectively 330.000, 10.000 and 10.000 data points. These sets were used as training, validation and test sets.

\section{Results}
\label{sec:result}
We were interested in comparing the amount of CF for networks trained with the \textit{Fine Tuning} procedure versus with the \textit{Gradual Tuning} procedure. In the first section (\ref{sec:Single task}) we present all the experiments where the Task-A is a classification performed on a single dataset. We tested the two procedures in the cases where Task-A and Task-B are rather simple, when Task-A and Task-B are more complex but similar, and when Task-A and Task-B are more complex and rather different.
In section \ref{sec:Multiple tasks} we study the amount of CF for a network trained on a task for which we suppose the network has good features. To guarantee this, Task-A was actually composed of 5 binary classification tasks on which the network was trained at the same time. The network was then trained on Task-B. In the first experiment Task-B is very similar to one of the tasks the network was trained on during Task-A. In the second experiment instead the classification task of Task-B is quite different from the classification tasks of Task-A. The results of all the experiments presented in paper are reported as the percentage of misclassified data on the test set. Frequently in the next paragraph we will refer to the error on the validation set, this refers to the percentage of misclassified data on the validation set.

\subsection{Single task experiment}
\label{sec:Single task}
The first experiment was on MNIST. During Task-A a network was trained on MNIST-04. The training was stopped if the error on the validation set did not go below its minimum for 20 consecutive epochs. The percentage of misclassification on the test set was reported. The training was done using a fixed learning rate $\lambda=0.01$. Once the training was over, the output of the last hidden layer of the network was connected to five extra output neurons for the classification on MNIST-59.

\begin{table}[t]
\caption{MNIST-04 $\rightarrow$ MNIST-59 (500-500). In the column named Task-B the mean percentage of misclassified on the test set calculated over 5 repetitions is reported together with its standard deviation. In the column named Task-A the first error reported for each entry is the percentage of misclassified cases on Task-A after the training on that task. The second and third value are the mean and the standard deviation of the percentage of misclassified data on Task-A after the training on Task-B. Column Epochs gives the mean and standard deviation of the number of epochs the network was trained on Task-B.} 
\begin{center}
    \begin{tabular}{| p{0.7cm} | p{1.7cm}| p{2.3cm} | p{1.1cm} | p{2cm} | p{2.6cm} | p{1.2cm} |}  
    \cline{2-7}   
    \multicolumn{1}{c|}{} & \multicolumn{3}{c|}{Fine Tuning} & \multicolumn{3}{c|}{Gradual Tuning}\\\hline
    \textbf{Reg}      & \textbf{Task-B}  & \textbf{Task-A} & \textbf{Epochs} & \textbf{Task-B} & \textbf{Task-A} & \textbf{Epochs} \\\hline
    No       & $1.56 \pm 0.07$ & $1.2 \rightarrow 4.5 \pm 1.2$ & $91 \pm 14$ & $1.54 \pm 0.05$ & $1.2 \rightarrow 1.83 \pm 0.05$ & $77 \pm 13$\\ \hline
    L1       & $1.51 \pm 0.03$ & $1.2 \rightarrow 3.5 \pm 0.6$ & $79 \pm 21$ & $1.44 \pm 0.03$ & $1.2 \rightarrow 2.02 \pm 0.02$ & $95 \pm 7$\\ \hline
    Drop	 & $1.60 \pm 0.07$ & $1.3 \rightarrow 8.2 \pm 0.9$ & $61 \pm 11$ & $1.41 \pm 0.06 $ & $1.3 \rightarrow 4.4 \pm 1.1$ & $113 \pm 38$\\ \hline
    \end{tabular}
\end{center}
\label{tab:Mnist-500}
\end{table}
The output layer parameters were randomly initialized. Training on Task-B was executed with both the \textit{Fine Tuning} procedure and the \textit{Gradual Tuning} procedure. In both cases the training was stopped when the error on the validation didn't go below its minimum in 20 epochs. Moreover in the \textit{Gradual Tuning} procedure, the training on a lower layer was allowed when the difference in the error on the validation set between two consecutive epochs was less than $0.1\%$. The network was trained on Task-B five times for each of the two training procedures, each time re-initializing the output layer. The entire experiment, training on Task-A and then on Task-B was executed using different regularization techniques. In particular we tried to train the network without any regularization, with L1, and with Dropout regularization. In tables \ref{tab:Mnist-500} and \ref{tab:Mnist-1000} the results relative to this experiment are reported for the two network architectures. In particular in the section of the table named Fine Tuning we report the results relative to the training with the \textit{Fine Tuning} procedure, while in the part of the table named Gradual Tuning we report the results obtained with the \textit{Gradual Training} technique. The column labeled Task-B contains the error on the test set for Task-B. The value reported in the table is the mean error obtained over the five repetitions and its standard deviation. Each entry of the column Task-A contains two error values. They refer to the error of the network on Task-A at the end of the training on Task-A and at the end of the training on Task-B. This second value is actually the mean error obtained over the five repetitions and it is reported with its standard deviation. Finally in the column labeled Epochs, we report the mean number of epoch the network was trained on Task-B with its standard deviation.\\
Looking at the two tables it is evident that the network trained with the \textit{Fine Tuning} procedure tends to forget much more than the one trained with the \textit{Gradual Tuning} procedure.
If we compare the error on Task-A after training on Task-B, we notice that the percentage of misclassifications in the case of \textit{Gradual Tuning} is generally around 2 times less than the percentage obtained with the \textit{Fine Tuning}. This happens despite the fact that in many cases the network trained with the \textit{Gradual Tuning} procedure is trained for more epochs than the one trained with the \textit{Fine Tuning} procedure. At the same time the network tends to perform similarly on the new task (Task-B) independently of the training procedure used. Finally it is also clear that the network trained with the Dropout regularization tends to forget much more than the network trained with L1 regularization or the one trained without regularization.
\begin{table}[h]
\caption{MNIST-04 $\rightarrow$ MNIST-59 (1000-1000-1000). See Table \ref{tab:Mnist-500} for a detailed presentation of the values reported.} 
\begin{center}
  \begin{tabular}{| p{0.7cm} | p{1.7cm}| p{2.35cm} | p{1.2cm} | p{2cm} | p{2.6cm} | p{1.2cm} |}  
    \cline{2-7}   
    \multicolumn{1}{c|}{} & \multicolumn{3}{c|}{Fine Tuning} & \multicolumn{3}{|c|}{Gradual Tuning}\\
    \hline
    \textbf{Reg}      & \textbf{Task-B}  & \textbf{Task-A} & \textbf{Epochs} & \textbf{Task-B} & \textbf{Task-A} & \textbf{Epochs} \\\hline
    No       & $2.01 \pm 0.03$ & $1.5 \rightarrow 4.3 \pm 0.9$ & $117 \pm 70$ & $1.99 \pm 0.05$ & $1.5 \rightarrow 2.6 \pm 0.1$ & $147 \pm 39$\\ \hline
    L1       & $1.92 \pm 0.05$ & $1.6 \rightarrow 3.4 \pm 1.3$ & $188 \pm 54$ & $1.88 \pm 0.02$ & $1.6 \rightarrow 2.0 \pm 0.1$ & $245 \pm 59$\\ \hline
    Drop     & $1.66 \pm 0.05$ & $1.3 \rightarrow 12.1 \pm 0.6$ & $102 \pm 17$ & $1.52 \pm 0.05$ & $1.3 \rightarrow 3.9 \pm 0.3$ & $138 \pm 35$\\ \hline
    \end{tabular}
\end{center}
\label{tab:Mnist-1000}
\end{table}
In order to verify the results obtained on MNIST on more complex datasets we did two other experiments using the synthetic datasets. In the first experiment Task-A and Task-B are very similar, while in the second the two tasks are more different. In the first experiment we used as Task-A the classification of the dataset \textit{acl} and as Task-B the classification on \textit{ac}. These are very similar tasks, with the first being more complex than the second because of the presence of a distracting line in the pictures. The tests were executed on the two network architectures and the results are reported in tables \ref{tab:acl-ac500}, \ref{tab:acl-ac1000}.
\begin{table}[h]
\caption{acl $\rightarrow$ ac (500-500). See Table \ref{tab:Mnist-500} for a detailed presentation of the values reported.} 
\begin{center}
\begin{tabular}{| p{0.7cm} | p{1.7cm}| p{2.35cm} | p{1.2cm} | p{2cm} | p{2.6cm} | p{1.2cm} |}  
    \cline{2-7}   
    \multicolumn{1}{c|}{} & \multicolumn{3}{c|}{Fine Tuning} & \multicolumn{3}{|c|}{Gradual Tuning}\\
    \hline
    \textbf{Reg}      & \textbf{Task-B}  & \textbf{Task-A} & \textbf{Epochs} & \textbf{Task-B} & \textbf{Task-A} & \textbf{Epochs} \\\hline
    No   & $2.50 \pm 0.01$ & $9.0 \rightarrow 18.8 \pm 0.3$ & $47 \pm 5$ & $2.50 \pm 0.05$ & $9.0 \rightarrow 16.1 \pm 0.3$ & $92 \pm 37$ \\ \hline
    L1   & $2.10 \pm 0.10$ & $7.1 \rightarrow 22.6 \pm 1.7$ & $38 \pm 7$ & $2.11 \pm 0.16$ & $7.1 \rightarrow 17.1 \pm 2.4$ & $45 \pm 23$\\ \hline
    Drop & $3.87 \pm 0.18$ & $7.7 \rightarrow 16.7 \pm 1.3$ & $115 \pm 40$ & $3.76 \pm 0.12$ & $7.7 \rightarrow 17.5 \pm 0.5$ & $151 \pm 17$\\ \hline
    \end{tabular}
\end{center}
\label{tab:acl-ac500}
\end{table}
In this case the difference between the \textit{Fine Tuning} and the \textit{Gradual Tuning} is less evident. The network performs almost the same on Task-B, but this time the amount of forgetting on Task-A is also nearly the same for the two procedures. The only exception is the big network (1000-1000-1000) when trained with L1 regularization. In that case we have again that the amount of forgetting is almost twice for \textit{Fine Tuning} as compared to \textit{Gradual Tuning}. 
\begin{table}[h]
\caption{acl $\rightarrow$ ac (1000-1000-1000). See Table \ref{tab:Mnist-500} for a detailed presentation of the values reported.} 
\begin{center}
\begin{tabular}{| p{0.7cm} | p{1.7cm}| p{2.35cm} | p{1.2cm} | p{2cm} | p{2.6cm} | p{1.2cm} |}  
    \cline{2-7}   
    \multicolumn{1}{c|}{} & \multicolumn{3}{c|}{Fine Tuning} & \multicolumn{3}{|c|}{Gradual Tuning}\\
    \hline
    \textbf{Reg}      & \textbf{Task-B}  & \textbf{Task-A} & \textbf{Epochs} & \textbf{Task-B} & \textbf{Task-A} & \textbf{Epochs} \\\hline
    No & $2.25 \pm 0.04$ & $6.7 \rightarrow 13.3 \pm 0.2$ & $73 \pm 52$ & $2.33 \pm 0.09$ & $6.7 \rightarrow 9.5 \pm 0.3$ & $47 \pm 16$ \\ \hline
    L1 & $1.60 \pm 0.05$ & $4.7 \rightarrow 20.5 \pm 2.9$ & $23 \pm 13$ & $1.61 \pm 0.06$ & $4.7 \rightarrow 11.9 \pm 3.2$ & $37 \pm 15$  \\ \hline
    Drop & $1.58 \pm 0.10$ & $5.9 \rightarrow 20.5 \pm 1.1$ & $113 \pm 21$ & $1.55 \pm 0.18$ & $5.9 \rightarrow 20.0 \pm 1.5$ & $145 \pm 46$ \\ \hline
    \end{tabular}
\end{center}
\label{tab:acl-ac1000}
\end{table}
The results presented in tables \ref{tab:acl-atl500} and \ref{tab:acl-atl1000} are relative to a network that was first trained on the dataset \textit{acl} (Task-A) and then on \textit{atl} (Task-B). Task-A is the same as for the previous experiment. Task-B is more complex when compared to Task-B of the previous experiment. Moreover in this experiment Task-A and Task-B are much less similar than in the previous experiment. This time the amount of forgetting seems to be different for the big network when trained without regularization or with L1 regularization (see Table \ref{tab:acl-atl1000}). In these two cases the network trained with \textit{Gradual Tuning} forgets approximately half as much as the network trained with \textit{Fine Tuning}. No particular difference seems to be present in the small network.
\begin{table}[h]
\caption{acl $\rightarrow$ atl (500-500). See Table \ref{tab:Mnist-500} for a detailed presentation of the values reported.} 
\begin{center}
\begin{tabular}{| p{0.7cm} | p{1.7cm}| p{2.35cm} | p{1.2cm} | p{2cm} | p{2.6cm} | p{1.2cm} |}  
    \cline{2-7}   
    \multicolumn{1}{c|}{} & \multicolumn{3}{c|}{Fine Tuning} & \multicolumn{3}{|c|}{Gradual Tuning}\\
    \hline
    \textbf{Reg}      & \textbf{Task-B}  & \textbf{Task-A} & \textbf{Epochs} & \textbf{Task-B} & \textbf{Task-A} & \textbf{Epochs} \\\hline
    No & $4.31 \pm 0.12$ & $9.1 \rightarrow 27.1 \pm 0.3$ & $78 \pm 16$ & $4.65 \pm 0.12$ & $9.1 \rightarrow 24.1 \pm 0.1$ & $120 \pm 28$\\ \hline
    L1 & $2.32 \pm 0.33$ & $7.1 \rightarrow 40.2 \pm 3.7$ & $77 \pm 48$ & $2.09 \pm 0.28$ & $7.1 \rightarrow 36.6 \pm 1.9$ & $108 \pm 20$\\ \hline
    Drop & $8.21 \pm 1.10$ & $7.7 \rightarrow 38.0 \pm 2.6$ & $158 \pm 46$ & $4.29 \pm 0.14$ & $7.7 \rightarrow 46.2 \pm 0.2$ & $293 \pm 4$\\ \hline
    \end{tabular}
\end{center}
\label{tab:acl-atl500}
\end{table}
\\
Considering all the experiments we described so far we found that the network trained with dropout tends to suffer more from CF than the one trained with L1 regularization, both for \textit{Fine Tuning} and \textit{Gradual Tuning} procedures. Moreover we often find that performance on the second task (Task-B) is often similar for the two regularization techniques. This may be due to the fact that the networks were trained with a fixed learning rate, while often dropout regularization is used either with a fixed learning rate, but bigger than the one we used, or with update of the weights with momentum. Such solutions were not adopted here because we wanted to use the same training procedure across all the experiments. 
\begin{table}[h]
\caption{acl $\rightarrow$  atl (1000-1000-1000). See Table \ref{tab:Mnist-500} for a detailed presentation of the values reported.} 
\begin{center}
\begin{tabular}{| p{0.7cm} | p{1.7cm}| p{2.35cm} | p{1.2cm} | p{2cm} | p{2.6cm} | p{1.2cm} |}  
    \cline{2-7}   
    \multicolumn{1}{c|}{} & \multicolumn{3}{c|}{Fine Tuning} & \multicolumn{3}{|c|}{Gradual Tuning}\\
    \hline
    \textbf{Reg}      & \textbf{Task-B}  & \textbf{Task-A} & \textbf{Epochs} & \textbf{Task-B} & \textbf{Task-A} & \textbf{Epochs} \\\hline
    No & $3.96 \pm 0.09$ & $6.7 \rightarrow 25.1 \pm 0.8$ & $120 \pm 56$ & $5.92 \pm 0.07$ & $6.7 \rightarrow 12.0 \pm 0.1$ & $191 \pm 12$\\ \hline
    L1 & $1.68 \pm 0.17$ & $4.7 \rightarrow 40.0 \pm 1.0$ & $56 \pm 9$ & $1.50 \pm 0.02$ & $4.7 \rightarrow 23.4 \pm 1.9$ & $56 \pm 6$\\ \hline
    Drop & $1.95 \pm 0.34$ & $5.9 \rightarrow 48.6 \pm 0.2$ & $167 \pm 57$ & $1.28 \pm 0.17$ & $5.9 \rightarrow 39.8 \pm 0.6$ & $205 \pm 57$\\ \hline
    \end{tabular}
\end{center}
\label{tab:acl-atl1000}
\end{table}

\subsection{Multiple tasks}
\label{sec:Multiple tasks}
In this section we describe the experiments for which Task-A consisted in training the network on 
several tasks at the same time. It is well known that a network trained at the same time on a set of 
similar tasks develops better features for each of the tasks than a network trained on just one of the tasks. 
This is in part due to the fact the network trained on several tasks is exposed to more data and 
in part to the ability of the network to transfer learning from the different tasks. So we were 
curious to see how much a network trained on several tasks would suffer from CF, and how the  
performance would differ for the \textit{Gradual Tuning} and the \textit{Fine Tuning} procedures. In the two 
experiments, during Task-A the network was trained on 5 datasets at the same time. The tasks 
we chose were the following: \textit{acl}, \textit{sb2l}, \textit{sbl}, \textit{sbt}, \textit{cnc}. The 
network we used for this experiment was a network of three hidden layers, each layer having 1000 
nodes. The last layer was connected to five output layers each one having two nodes. During the 
training, batches of 20 data points from the different datasets were alternated in a fixed order. At the end of each epoch 
the error on the validation set was calculated for all of the five datasets. If all of the errors on the 
validation were below their minimum the network configuration was stored. If any of the 5 errors did 
not go below its minimum in 20 epochs the training was stopped. Once the training was over a sixth output 
layer was connected to the last hidden layer and the network was trained on Task-B. The training on Task-B was repeated 5 times in the two conditions (\textit{Fine Tuning} and \textit{Gradual Tuning}) each time by randomly 
initializing the new output layer. In the experiments Task-B always consisted of a single 
binary classification task on one of the datasets described in \ref{sec:tasks}.\\
\begin{table}
\caption{acl + sb2l + sbl + sbt + cnc $\rightarrow$ ac (1000-1000-1000). The results are reported following the same schema as in table \ref{tab:acl-ac500}, but this time Task-A consisted in training the network on 5 datasets. Each row in column Task-A refers to one of the datasets. The three numbers reported are respectively the error on the task at the end of Task-A, the mean error and its standard deviation at the end of Task-B over 5 repetitions.}
\begin{center}
\begin{tabular}{|p{0.6cm}|p{1.55cm}|l|p{1.25cm}|p{1.55cm}|l|p{1.25cm}|}
\cline{2-7}   
    \multicolumn{1}{c|}{} & \multicolumn{3}{c|}{Fine Tuning} & \multicolumn{3}{c|}{Gradual Tuning}\\\hline
    \textbf{Reg}      & \textbf{Task-B}  & \textbf{Task-A} & \textbf{Epochs} & \textbf{Task-B} & \textbf{Task-A} & \textbf{Epochs} \\\hline
\multirow{5}{*}{No} & \multirow{5}{*}{$ 1.66 \pm 0.06 $} & $ 7.0 \rightarrow 11.6 \pm 0.4 $ (acl)& \multirow{5}{*}{$ 42 \pm 29 $} & \multirow{5}{*}{$ 1.82 \pm 0.08 $} & $ 7.0 \rightarrow 10.2 \pm 0.3 $ & \multirow{5}{*}{$ 98 \pm 54 $}\\
& & $5.5 \rightarrow 7.4 \pm 0.2$ (sb2l) & & & $5.5 \rightarrow 6.6 \pm 0.3$ & \\
& & $2.4 \rightarrow 3.9 \pm 0.4$ (sbl)  & & & $2.4 \rightarrow 2.8 \pm 0.1$ & \\
& & $1.5 \rightarrow 2.3 \pm 0.1$ (sbt)  & & & $1.5 \rightarrow 1.8 \pm 0.1$ & \\
& & $0.9 \rightarrow 1.2 \pm 0.1$ (cnc)  & & & $0.9 \rightarrow 1.1 \pm 0.1$ & \\ \hline
\multirow{5}{*}{L1} & \multirow{5}{*}{$ 0.92 \pm 0.02 $} & $ 4.4 \rightarrow 21.0 \pm 1.0 $ (acl)& \multirow{5}{*}{$ 26 \pm 5 $} & \multirow{5}{*}{$ 0.89 \pm 0.04 $} & $ 4.4 \rightarrow 14.8 \pm 0.8 $ & \multirow{5}{*}{$ 36 \pm 5 $}\\
& & $3.3 \rightarrow 6.3 \pm 0.5$ (sb2l)& & & $3.3 \rightarrow 5.4 \pm 0.3$ & \\
& & $1.4 \rightarrow 5.6 \pm 0.5$ (sbl) & & & $1.4 \rightarrow 2.1 \pm 0.0$ & \\
& & $0.7 \rightarrow 2.6 \pm 0.4$ (sbt) & & & $0.7 \rightarrow 1.6 \pm 0.1$ & \\
& & $0.4 \rightarrow 0.7 \pm 0.1$ (cnc) & & & $0.4 \rightarrow 0.8 \pm 0.0$ & \\ \hline
\multirow{5}{*}{Drop} & \multirow{5}{*}{$ 0.73 \pm 0.08 $} & $ 4.1 \rightarrow 11.5 \pm 1.6 $ (acl) & \multirow{5}{*}{$ 58 \pm 22 $} & \multirow{5}{*}{$ 0.70 \pm 0.10 $} & $ 4.1 \rightarrow 11.6 \pm 1.7 $ & \multirow{5}{*}{$ 88 \pm 23 $}\\
& & $2.4 \rightarrow 6.0 \pm 0.7$ (sb2l)& & & $2.4 \rightarrow 6.1 \pm 0.6$ & \\
& & $0.8 \rightarrow 4.1 \pm 0.7$ (sbl) & & & $0.8 \rightarrow 4.1 \pm 0.7$ & \\
& & $0.2 \rightarrow 1.0 \pm 0.2$ (sbt) & & & $0.2 \rightarrow 1.1 \pm 0.2$ & \\
& & $0.4 \rightarrow 0.5 \pm 0.0$ (cnc) & & & $0.4 \rightarrow 0.6 \pm 0.1$ & \\ \hline
\end{tabular}
\end{center}
\label{tab:5task1}
\end{table} 
In the first experiment we evaluated the amount of CF in the case where Task-B is a simplified 
version of one of the tasks proposed in Task-A; we used as Task-B the \textit{ac} task. The results are reported in 
table \ref{tab:5task1}. The results are presented in the table following the same schema as in the 
previous table, the only difference consists in the fact that the column named Task-A now has the results relative to the five datasets before and 
after the training on Task-B. We can now compare the amount of forgetting on \textit{acl}
and the performance on \textit{ac} obtained in this experiment with the performance obtained in the 
experiment described in the previous section where the network was trained first on 
\textit{acl} and then on \textit{ac}(see table \ref{tab:acl-ac1000}). It is evident that the 
performance on \textit{ac} is much better in this case than before. Regarding the amount of CF that 
the two networks suffer, the results seem very similar and it is not easy to say whether CF for the network that has been trained on several tasks actually is less strong. The effect of the \textit{Gradual Tuning} in this case is also quite similar to the effect we obtained on the network trained on a
single task. As before we observe that the network trained with the \textit{Gradual Tuning} procedure has very 
similar performances on the new task when compared with the network trained with the \textit{Fine Tuning} 
procedure. Also the network trained with the \textit{Gradual Tuning} procedure tends to forget less than the one trained normally, although the improvement in this case is less evident than it was in other cases.\\
In this second experiment Task-B is very different from any of the classifications performed during Task-A because we chose Task-B to be the \textit{atl} task. The results are reported in table \ref{tab:5task2}. This experiment is similar to the last experiment of the previous section. In that case we trained a network first on \textit{acl} and then on \textit{atl}(see table \ref{tab:acl-atl1000}). 
\begin{table}
\caption{acl + sb2l + sbl + sbt + cnc $\rightarrow$ atl (1000-1000-1000). See Table \ref{tab:5task1} for a detailed presentation of the values reported.}
\begin{center}
\begin{tabular}{|p{0.6cm}|p{1.55cm}|l|p{1.25cm}|p{1.55cm}|l|p{1.25cm}|}
\cline{2-7}   
    \multicolumn{1}{c|}{} & \multicolumn{3}{c|}{Fine Tuning} & \multicolumn{3}{c|}{Gradual Tuning}\\\hline
    \textbf{Reg}      & \textbf{Task-B}  & \textbf{Task-A} & \textbf{Epochs} & \textbf{Task-B} & \textbf{Task-A} & \textbf{Epochs} \\\hline
\multirow{5}{*}{No} & \multirow{5}{*}{$ 2.68 \pm 0.11 $} & $ 7.0 \rightarrow 17.6 \pm 0.7 $ (acl) & \multirow{5}{*}{$ 116 \pm 66 $} & \multirow{5}{*}{$ 3.46 \pm 0.12 $} & $ 7.0 \rightarrow 11.8 \pm 0.1 $ & \multirow{5}{*}{$ 153 \pm 59 $}\\
& & $5.5 \rightarrow 16.7 \pm 0.9$ (sb2l) & & & $5.5 \rightarrow 11.1 \pm 0.1$ & \\
& & $2.4 \rightarrow 21.8 \pm 2.1$ (sbl)& & & $2.4 \rightarrow 9.1 \pm 0.1$ & \\
& & $1.5 \rightarrow 4.2 \pm 0.2$  (sbt)& & & $1.5 \rightarrow 2.6 \pm 0.1$ & \\
& & $0.9 \rightarrow 7.1 \pm 0.7$  (cnc)& & & $0.9 \rightarrow 1.5 \pm 0.1$ & \\ \hline
\multirow{5}{*}{L1} & \multirow{5}{*}{$ 1.29 \pm 0.06 $} & $ 4.4 \rightarrow 17.8 \pm 4.0 $ (acl) & \multirow{5}{*}{$ 17 \pm 2 $} & \multirow{5}{*}{$ 1.14 \pm 0.01 $} & $ 4.4 \rightarrow 11.3 \pm 0.6 $ & \multirow{5}{*}{$ 33 \pm 5 $}\\
& & $3.3 \rightarrow 12.6 \pm 1.2$ (sb2l)& & & $3.3 \rightarrow 9.0 \pm 0.2$ & \\
& & $1.4 \rightarrow 7.5 \pm 1.3$  (sbl) & & & $1.4 \rightarrow 5.3 \pm 0.4$ & \\
& & $0.7 \rightarrow 6.5 \pm 2.4$  (sbt) & & & $0.7 \rightarrow 1.5 \pm 0.1$ & \\
& & $0.4 \rightarrow 19.3 \pm 4.8$ (cnc) & & & $0.4 \rightarrow 5.1 \pm 0.6$ & \\ \hline
\multirow{5}{*}{Drop} & \multirow{5}{*}{$ 0.59 \pm 0.09 $} & $ 4.1 \rightarrow 12.4 \pm 1.9 $ (acl) & \multirow{5}{*}{$ 95 \pm 40 $} & \multirow{5}{*}{$ 0.52 \pm 0.06 $} & $ 4.1 \rightarrow 10.1 \pm 0.7 $ & \multirow{5}{*}{$ 138 \pm 22 $}\\
& & $2.4 \rightarrow 9.5 \pm 1.1$ (sb2l)& & & $2.4 \rightarrow 8.5 \pm 0.5$ & \\
& & $0.8 \rightarrow 8.0 \pm 0.3$ (sbl)& & & $0.8 \rightarrow 5.0 \pm 0.4$ & \\
& & $0.2 \rightarrow 3.0 \pm 0.5$ (sbt)& & & $0.2 \rightarrow 2.0 \pm 0.1$ & \\
& & $0.4 \rightarrow 29.6 \pm 6.2$(cnc)& & & $0.4 \rightarrow 17.6 \pm 2.5$ & \\ \hline
\end{tabular}
\end{center}
\label{tab:5task2}
\end{table}
The results on Task-B are much better for the network that was trained on several tasks during Task-A, when 
compared to the network that was trained just on one task \ref{tab:acl-atl1000}). Concerning the amount of CF on
task \textit{acl} two considerations are interesting. First, this network forgets much 
less than the network trained on a single task during Task-A. This suggests that having better features 
helped reducing CF. 
Regarding the second consideration let's compare the results in the tables \ref{tab:5task1} and 
\ref{tab:5task2}. These were obtained with the same Task-A. The amount of forgetting on \textit{acl} is 
much less in the experiment reported in table \ref{tab:5task1}, where the new task is very similar to 
the old one. Lastly the \textit{Gradual Tuning} technique seems to be quite effective in this case, the amount of CF being reduced in all the cases, most of the time being from 1.5 to 5 times less. 

\section{Discussion}
The \textit{Gradual Tuning} technique proved to be a good strategy for reducing the amount of CF in a transfer learning task. The CF was smaller in almost all the experiments we described in the paper. Regarding the performance on a new task, we can say that the networks trained with the two procedures showed similar performance in all the experiments except in the MNIST experiment, where the networks trained with \textit{Gradual Tuning} 
showed better results. Although we didn't formally verify it, it is most probably the case that compared to \textit{Fine Tuning}, \textit{Gradual Tuning} tends to preserve more of the features of the original task. For this reason we would have expected that, at least when the two tasks are very similar, the network trained with \textit{Gradual Tuning}, in addition to forgetting less on the old task, would also outperform the network trained with \textit{Fine Tuning} on the new task. Further experiments are necessary to understand why this was not the case.

\subsubsection*{Acknowledgments}
This work was funded by ERC Advanced Grant 323674 “FEEL” and ERC proof of concept grant 692765 "FeelSpeech" to Kevin O'Regan.

\newpage
\bibliographystyle{plain}

\end{document}